\documentclass[conference]{IEEEtran}
\usepackage{cite}
\usepackage{amsmath,amssymb,amsfonts}
\usepackage{graphicx}
\graphicspath{ {./images/} }
\usepackage{textcomp}
\usepackage{adjustbox}
\usepackage[section]{placeins}
\usepackage{longtable}
\usepackage[utf8]{inputenc}
\usepackage{array}
\usepackage{multirow}
\usepackage{comment}
\usepackage{wrapfig}
\usepackage{float}
\usepackage{caption}
\usepackage[colorlinks = true,
            linkcolor = blue,
            urlcolor  = blue,
            citecolor = blue,
            anchorcolor = blue]{hyperref}
\usepackage{url}

\usepackage{caption}
\usepackage{subcaption}

\usepackage{color}
\usepackage{booktabs}

\usepackage{xcolor}
\usepackage[linesnumbered,ruled,vlined]{algorithm2e}
\usepackage{amsmath}

\SetCommentSty{mycommfont}
\SetKwInput{KwInput}{Input}
\SetKwInput{KwOutput}{Output}

\begin{document}

\title{Optimizing Fintech Marketing: A Comparative Study of Logistic Regression and XGBoost}

\author{
\IEEEauthorblockN{Dinesh Chowdary Attota}
    \IEEEauthorblockA{
    \textit{Department of Data Science \& Analytics}\\
     \textit{College of Computer Science and Software Engineering}\\
        \textit{Kennesaw State University}\\
        Marietta, GA, USA \\
        \textit{dattota@students.kennesaw.edu}}
\and
\IEEEauthorblockN{Sahar Yarmohammadtoosky }
 \IEEEauthorblockA{
    \textit{Department of Data Science \& Analytics}\\
     \textit{College of Computer Science and Software Engineering}\\
        \textit{Kennesaw State University}\\
        Marietta, GA, USA \\
\IEEEauthorblockA{\textit{syarmoha@students.kennesaw.edu} \\
}
}
}

\maketitle

\begin{abstract}
As several studies have shown, predicting credit risk is still a major concern for the financial services industry and is receiving a lot of scholarly interest. This area of study is crucial because it aids financial organizations in determining the probability that borrowers would default, which has a direct bearing on lending choices and risk management tactics. Despite the progress made in this domain, there is still a substantial knowledge gap concerning consumer actions that take place prior to the filing of credit card applications. The objective of this study is to predict customer responses to mail campaigns and assess the likelihood of default among those who engage. This research employs advanced machine learning techniques, specifically logistic regression and XGBoost, to analyze consumer behavior and predict responses to direct mail campaigns. By integrating different data preprocessing strategies, including imputation and binning, we enhance the robustness and accuracy of our predictive models. The results indicate that XGBoost consistently outperforms logistic regression across various metrics, particularly in scenarios using categorical binning and custom imputation. These findings suggest that XGBoost is particularly effective in handling complex data structures and provides a strong predictive capability in assessing credit risk.

\end{abstract}


\IEEEpeerreviewmaketitle
\section{Introduction}

An essential part of the financial services industry is credit risk analysis, which focuses on managing and evaluating possible lending-related risks. For financial institutions, determining an applicant's likelihood of repaying a loan is the most important factor in the decision to grant credit. Determining the right recipients for credit card offers is another strategic concern for lenders. This targeting is essential to reducing the risks posed by prospective defaults or non-responses from those with inadequate credit records. 

This article attempts to investigate these credit risk factors in depth. It aims to improve the effectiveness of financial product distribution and reduce financial losses by not only analyzing the inherent risks in credit distribution but also improving the tactics for guiding credit card marketing campaigns.
Large datasets with high dimensionality and complicated, frequently undefinable interactions between data variables provide analysts with many hurdles when developing predictive models. One common technique used in the financial services sector for binary classification, logistic regression, sometimes performs inadequately when there are a lot of variables. This constraint arises from its reduced effectiveness in handling extremely nonlinear interactions between these variables. As a result, other approaches to machine learning have become more popular. XGBoost is especially valuable for its improved computing efficiency and adaptability to big, complicated datasets with unknown features. Because of its strong performance in a range of predictive modelling scenarios and ability to effectively address the drawbacks of more conventional statistical methods, this technique is becoming increasingly common.
In this study, we develop Logistic Regression and XGBoost models to predict commercial credit risk and assess the likelihood of responses to a mail campaign using datasets provided by Atlanticus. Following a review of previous research, this paper outlines the data preprocessing steps in detail, including imputation, discretization, and multicollinearity assessment, which are employed to select variables from a large but sparse dataset. Subsequently, the methodologies of Logistic Regression and XGBoost are described. Finally, the paper discusses the results and compares the models based on various metrics such as accuracy, precision, recall, F1 score, and ROC curve analysis.

\section{Related Work} \label{sec:relatedwork}
Making credit decisions is an important financial decision involving many factors in a dynamic market. The risks associated with digital financial transactions are increasing along with their use \cite{almudaires2021data}transactions are increasing along with their use \cite{sym}. In order to predict delinquency, Almudaires, F. \cite{almudaires2021data} linked consumer tradeline, credit bureau, and macroeconomic factors using machine learning approaches. They discovered significant variation among banks in terms of risk variables, sensitivity, and predictability of delinquency. The enhanced Credit Card Risk Identification (CCRI) technique, as proposed by Rtayli, N. \cite{rtayli2020selection}, beat certain algorithms, such as decision trees, in terms of classification performance. It was based on the features selection algorithm as Random Forest Classifier and Support Vector Machine to detect fraud risk. 

XGBoost and Long-Short Term Memory (LSTM) were used for comparative analysis in this study \cite{shokrollahi2023comprehensive} \cite{gao2021research}. According to the study, the XGBoost model's performance is influenced by the level of feature extraction competence, whereas the LSTM method can reach higher accuracy even in the absence of feature extraction. Dwidarma, R. \cite{dwidarma2021comparison} estimated the likelihood that clients would accept credit facilities or enter debt. The study's comparison of the XGBoost approach with the logistic regression model demonstrated that the latter produces superior outcomes. 


\section{Data Discovery} \label{sec: Proposed Approach}
This large-scale study examined data related to direct mail campaigns from a Fintech company, and campaigns run roughly every 3 weeks. Data contains both non-responders and trade performance for those who did open a card at the specified time. It has about 988,000 observations, and 541 variables. This study focuses on specifying how likely it is for a customer to respond to the mail, and how likely it is that a booked customer will not default.  

\subsection{Dependent Variable Assignment}
\begin{figure*}
   \begin{minipage}{0.3\textwidth}
     \centering
     \includegraphics[width=0.8\linewidth,height=1.4in]{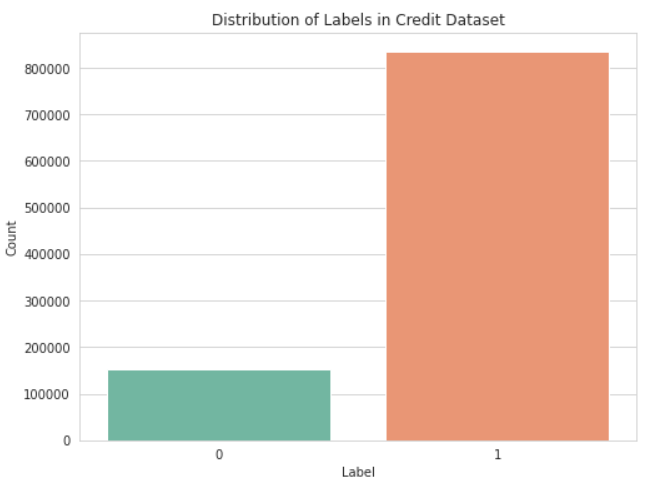}
    \caption{Mail or Don't Mail Model}
    \label{fig:mail_or_dont_mail_distribution}
   \end{minipage}\hfill
   \begin{minipage}{0.32\textwidth}
     \centering
    \includegraphics[width=.97\linewidth,height=1.4in]{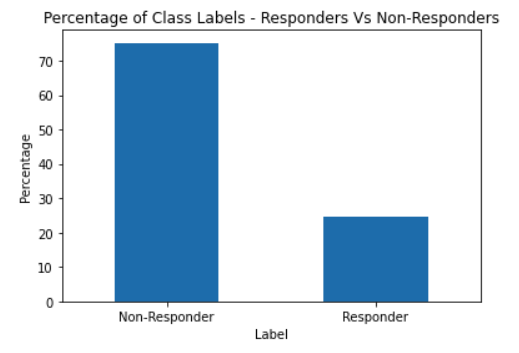}
    \caption{Responders Model}
    \label{fig:responders_model_label_distribution}
   \end{minipage}
   \begin{minipage}{0.32\textwidth}
     \centering
     \includegraphics[width=.97\linewidth,height=1.4in]{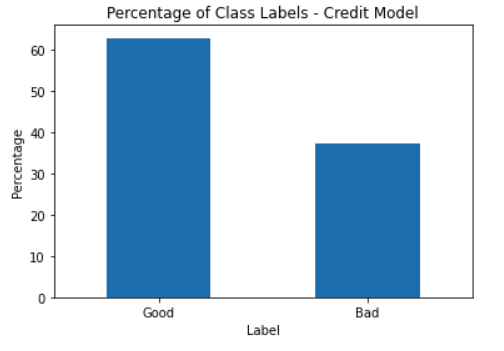}
    \caption{Credit Model}
    \label{fig:credit_model_label_distribution}
   \end{minipage}
\vspace{-0.1in}
\end{figure*}

This study aims to assess the effectiveness of targeted postal marketing in forecasting consumer responses and the corresponding credit risks. The main variable of interest, referred to as GoodBad, classifies customer actions after getting a credit card. More precisely, when the value of goodbad is 0, it means that a person has obtained a credit card and consistently made payments on time without any delinquency. On the other hand, when goodbad is 1, it shows that there has been delinquency after the credit card was opened. Instances in which GoodBad is absent are considered as non-responses to the mail campaign.
\begin{enumerate}
    \item \textbf{Mail or Don't Mail Model} : In order improve our understanding, we have devised an innovative modeling technique that assists in finding the optimal course of action about whether or not to send mail, therefore facilitating the decision-making process. This model classifies potential responses into two distinct outcomes: 'Mail' (label 1), indicating that the consumer is likely to respond to the campaign and is considered creditworthy, and 'Don't Mail' (label 0), encompassing those who are unlikely to respond or may have a higher credit risk. Figure \ref{fig:mail_or_dont_mail_distribution} depicts the distribution of the dependent variable employed in the mail or don't mail model.
    
    \item \textbf{Responders Model} : The second modeling approach focused on binary classification, with the goal of predicting the distinction between individuals who responded and those who did not respond to the mail campaign. In the 'Responders Model,' we created a new binary dependent variable called 'Responders.' It is coded as '1' for individuals who participated in the campaign and '0' for those who did not respond. The distribution of labels, shown in Figure \ref{fig:responders_model_label_distribution}, reveals that the majority of the dataset, 75.2\%, falls under the category of 'Non-responders.' The remaining 24.79\% of the dataset is classified as 'Responders'.

    \item \textbf{Credit Model}: The third approach, referred to as the 'Credit Model,' utilized binary classification to primarily forecast consumer credit risk. By analyzing the dataset, we were able to distinguish between consumers with a positive credit standing and those who were at risk of default. As shown in Figure \ref{fig:credit_model_label_distribution}, our dataset for this model was made up of 62.72\% consumers labeled as 'good' and 37.27\% as 'bad.' We specifically focused on respondents who had opened trade with the Fintech company.
\end{enumerate} 

This methodology not only improves the accuracy of predictions but also offers a detailed understanding of how consumers respond to direct mail advertisements.

\subsection{Data Cleansing and Imputation}

During the data preparation phase, we took great care to thoroughly clean the dataset and implement systematic imputation procedures to guarantee high-quality data for effective model training. The cleaning and imputation process for independent variables involved a series of sequential steps:
\begin{figure*}
    \centering
    \includegraphics[scale=0.28]{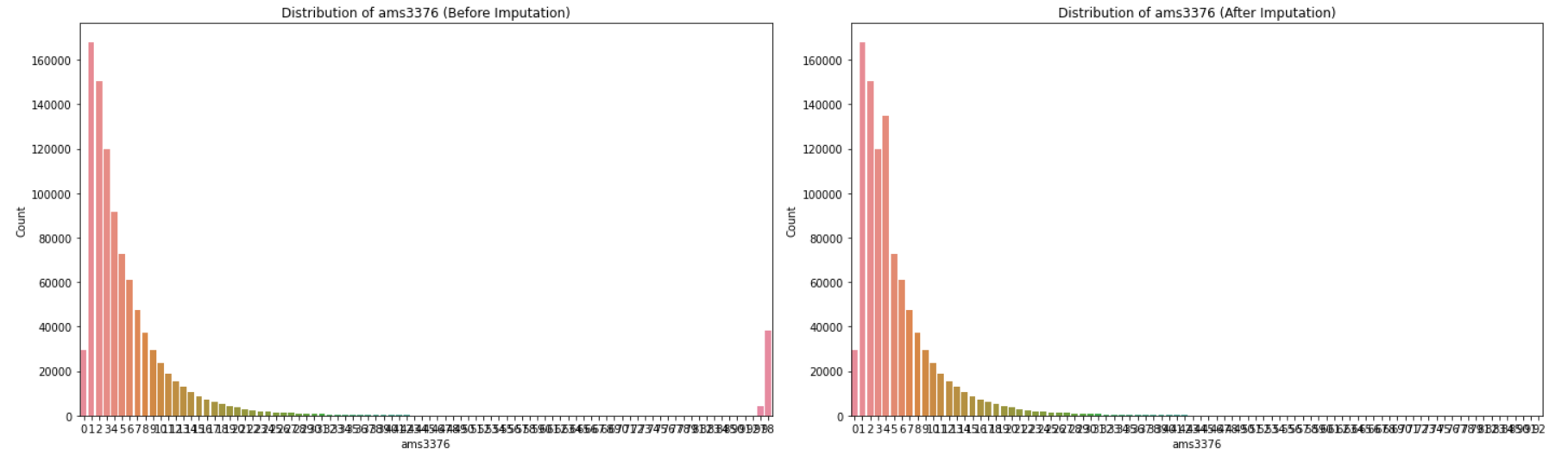}
    \caption{Example of Custom Bin Imputation}
    \label{fig:median_imputation}
\end{figure*}

\begin{figure*}
        \centering
        \includegraphics[scale=0.28]{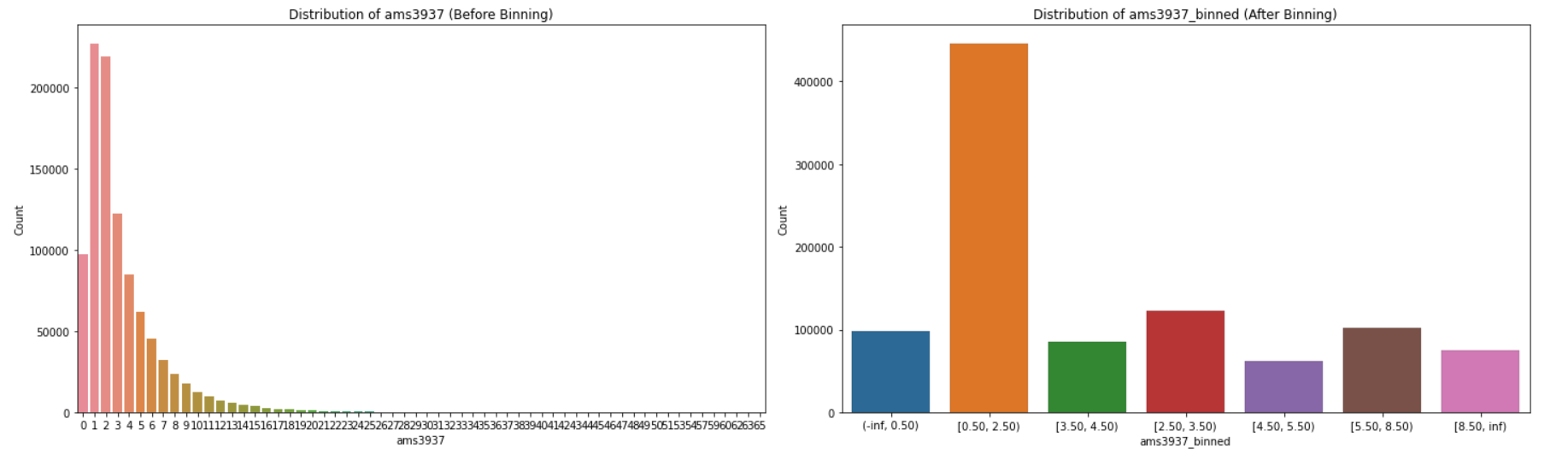}
        \caption{Example of Binning}
        \label{fig:binning}
    \end{figure*}

\begin{enumerate}
    \item \textbf{Step1 (Identification of Coded Values)}: We first discovered coded values within the dataset, specifically numerical values that start with '9' and end with any digit from '4' to '9', as long as the value has two or more digits. This rule was implemented for all independent variables to determine a complete range of coded values. After conducting our investigation, we were able to identify 17 unique coded values. These values are listed in Table \ref{tab:unique_coded_values} for reference.
    Also, the variable age(ams3746) was dropped from the dataset because credit worthiness cannot be determined using the age of the customers.
    \begin{table}[]
    \centering
    \begin{tabular}{|l|}
    \hline
    \textbf{Unique Coded Values}                                                                                                                                            \\ \hline
    \begin{tabular}[c]{@{}l@{}}96, 97,98,  99,  9444,  9996,\\  9997,  9998,  9999,  99994,\\  99996,  99997,  99998,\\  99999,  9999996,\\  9999997,  9999998\end{tabular} \\ \hline
    \end{tabular}
    \caption{Unique Coded Values in the Dataset}
    \label{tab:unique_coded_values}
    \end{table}
    
    As part of our analysis, we began by reclassifying specific coded values as missing data. The coded values, with their distinct numerical patterns, were found in all the independent variables. As a result, any column that had more than 30\% missing or coded values was removed from the analysis.

    \item \textbf{Step 2 (Elimination of Single-Valued Columns)}: In our next step, we focused on columns that had the same value across each row. After realizing that these columns don't provide any useful information for predictive modeling, we decided to remove them in order to improve the performance of the model. We found and removed a total of 9 columns.

    \item \textbf{Step 3 (Imputation)}: This work employs a dual imputation strategy, which involves using both median imputation and a unique bin imputation approach, to tackle the issues presented by missing data. The choice of median imputation was based on its ability to effectively handle skewed distributions and outliers. Unlike mean imputation, median imputation gives a reliable estimate of central tendency that is not affected by extreme values. This strategy guarantees the preservation of the true distribution of the dataset.
    This work addresses the difficulties caused by missing data by employing a dual imputation strategy, which involves using both median imputation and a proprietary bin imputation method. The choice of median imputation was based on its ability to handle skewed distributions and outliers effectively. Unlike mean imputation, median imputation provides a reliable measure of central tendency that is not affected by extreme values. This strategy guarantees the preservation of the true distribution of the dataset.

    In order to enhance the accuracy of our data handling methods, we have incorporated a specialized approach known as custom bin imputation. The concept is summarized as follows:

    \begin{enumerate}
        \item Data Segmentation: The data for the variable was divided into bins of 5
        \item Rate Calculation: We calculated the 'good rate' for each bin by determining the proportion of favorable outcomes, such as no delinquency, and also calculated the median number.
        \item Rate Matching: Next, we computed the rate of occurrence for each coded value within the variable.
        \item Value Assignment: Each encoded value was replaced with the mean value of the category whose rate of success most closely matched the rate of success of the encoded value.
        This method guarantees an accurate treatment of missing values, enhancing the accuracy of estimated values by closely matching them with the statistical properties of their corresponding groups. The Figure \ref{fig:binning} demonstrates how several imputation approaches affect the distribution of variables, especially when dealing with coded values and outliers.
    \end{enumerate}

\end{enumerate}

\subsection{Data Skewness and Binning}
In this section, we will explore the strategies employed to identify whether a column should be classified as continuous or discrete. We conducted a thorough analysis of the skewness and kurtosis values of the column distribution to determine whether the data should be classified as continuous or discrete.

\begin{enumerate}
    \item \textbf{Skewness and Kurtosis}: Our analytical framework involves a quantitative assessment of the symmetry and tailedness of the dataset's distribution. This is done by calculating skewness and kurtosis for each column, using robust statistical functions found in standard libraries. We set up thresholds to categorize the characteristics of these distributions.
    \begin{enumerate}
        \item \textbf{Kurtosis}: A column is considered to have "High Kurtosis" when its kurtosis value exceeds 20. This suggests that the distribution has heavier tails compared to a normal distribution. Columns that do not meet this threshold are identified as having "Low Kurtosis."

        \item \textbf{Skewness}: The metric for skewness helps us gain insight into the asymmetry of a column. If the skewness value is greater than 2 or less than -2, it indicates a "High Skewness," indicating a significant departure from a symmetrical distribution. Values falling within these limits are classified as "Low/Moderate Skewness".
        \item A column is considered "Discrete" if it has less than 15 unique values and shows both "High Skewness" and "High Kurtosis" at the same time. On the other hand, columns that do not meet these criteria at the same time are considered "Continuous." Understanding this distinction is crucial for the subsequent data processing and modeling. It helps in selecting the right statistical techniques and algorithms for each variable type.

    \end{enumerate}

    \item \textbf{Binning}: In our project, we successfully implemented an optimal binning strategy by utilizing the mathematical programming formulation for optimal binning, as outlined by Guillermo Navas-Palencia \cite{navaspalencia2022optimal}. This approach offers a systematic way to categorize variables into bins, addressing concerns like extreme values, and complex associations. As therefore, it improves the understandability and effectiveness of the model. The efficiency of the system is derived from the utilization of constrained programming solvers. These solvers strive to identify the bins that have the most statistical significance while still meeting the limitations provided by the user. This allows for a reliable and adaptable method to binning. The OptimalBinning class utilizes parameters such as $dtype$ to indicate the variable type and $solver$ to select the optimization method, ensuring flexibility to accommodate different data features.
   A key component of our implementation involved calculating the Weight of Evidence (WoE) and Information Value (IV) for each variable after binning. comprehending the predictive power of binned variables is crucial in assessing their impact. Two important metrics, WoE and IV, provide valuable insights. WoE helps us understand the relationship between predictors and the target variable, while IV measures the overall predictive capability of a feature. 
   The Information Value (IV) measures how well a variable may be predicted, with weak, medium, and strong being the usual interpretations of its values. The IV provides insight into a variable's utility in class distinction by adding up the products of the difference in Goods and Bads distributions across bins and their WoE. For the initial testing round, we decided to use a threshold of IV $>$ 0.1 for variable identification and selection. This criterion ensured that only features with a medium to strong predictive power were kept, thus improving the efficiency and accuracy of our modeling efforts. The fig \ref{fig:binning} shows an example of how the continuous data of a column has been categorized into different bins.

    \begin{figure*}
    \centering
    \includegraphics[scale=0.15]{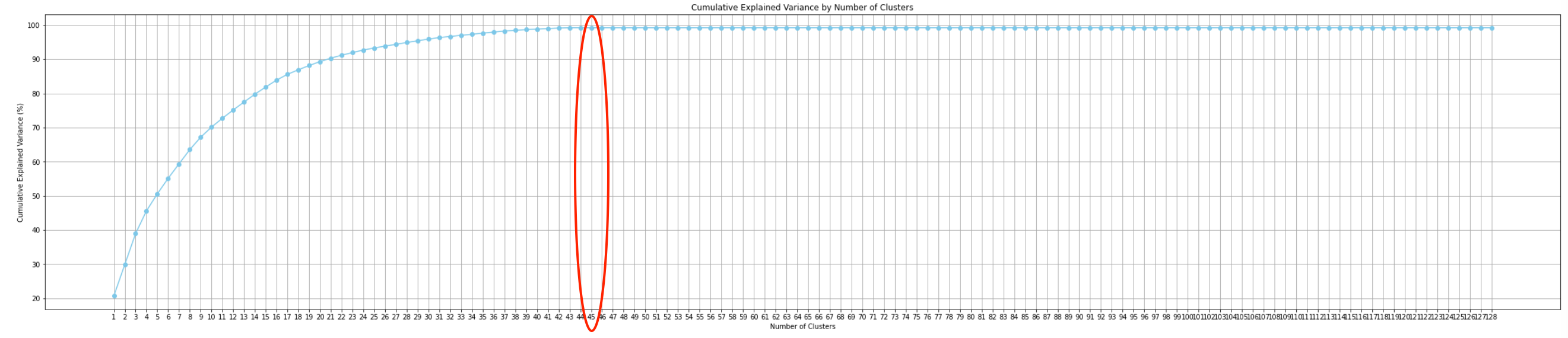}
    \caption{Proportion of Variance}
    \label{fig:PoV}
    \end{figure*}

   \item \textbf{Variable Clustering}: We implemented Variable Clustering, an innovative approach underpinned by three fundamental steps: Principal Component Analysis (PCA), Eigenvalues and Communalities, and the 1 – R\_Square Ratio. This methodology, adeptly applied using both SAS and Python, serves to distill the essence of our dataset, ensuring a robust yet simplified model input.
   \begin{enumerate}
       \item \textbf{Principal Component Analysis}: PCA is a fundamental component of our Variable Clustering algorithm, providing a valuable perspective on the correlation among variables. Through the use of a correlation matrix, PCA allows for the calculation of principal components (PCs), which are weighted linear combinations of predictor variables. This process simplifies the complex inter-variable relationships in the dataset, reducing them to a more manageable number of components.
        \item \textbf{Eigenvalues and Communalities}: The strength of each PC is measured by eigenvalues, which indicate the amount of variance that each PC explains in the dataset. On the other hand, communalities represent the total amount of variance that is accounted for through all principal components for a specific variable, adding up to a total of 100\%. This step helps identify the most important PCs, usually the initial ones, that capture the majority of data variation and are then used for further analysis.
        \item \textbf{1 – R\_Square Ratio}: The selection of the representative variable for each cluster depends on the 1 – R\_Square Ratio. This ratio helps guide the decision-making process by favoring variables that have a strong correlation within their own clusters and a weak correlation with variables in other clusters. This criterion ensures that the selected variable accurately reflects the cluster's characteristics, making it easier to reduce dimensions without losing significant information.
   \end{enumerate}

   \item \textbf{Proportion of Variance (PoV)}: In our study, we delved into the degree of predictive power within our dataset by conducting a Cumulative Proportion of Variance (PoV) analysis, which is a crucial aspect of Principal Component Analysis (PCA). The PoV graph, as shown in the Fig \ref{fig:PoV}, is an essential for understanding the number of variables needed to capture a significant portion of the dataset's variance.

    The graph illustrates the cumulative explained variance as we increase the number of variables (or principal components). Interestingly, the plateau of the curve suggests a stage where the addition of more variables has little impact on the overall explanatory power. Through our analysis, we have identified a crucial point where 99\% of the dataset's variance is captured, involving 45 variables. This discovery demonstrates the effectiveness of our variable selection process, showing that a significant amount of information can be represented by less than half of the original variables.

    \item \textbf{Variance Inflation Factor}: The selected subset of variables after variable clustering and proportion of variance calculation, underwent a Variance Inflation Factor (VIF) analysis to identify and address multicollinearity, thereby ensuring the stability and interpretability of the predictive model. We used a VIF threshold of 3, which is in line with standard statistical practices. Variables with a VIF value higher than 3 may indicate potential issues with multicollinearity. In order to efficiently address and resolve multicollinearity, we developed an iterative VIF reduction process. During each iteration, the VIF values for all variables were calculated and then sorted in descending order. A variable with a high VIF value that exceeded the threshold of 3 was eliminated from the dataset. This iterative process continued until all remaining variables showed VIF values below 3, ensuring that no single variable had a disproportionate impact on the variance explained by other variables.
    
\end{enumerate}


\section{Evaluation \& Results} \label{sec: Evaluation Results}
\subsection{Datasets}
The dataset used for this study was generously provided by a prominent fintech company that specializes in customised financial services, such as credit offerings. This dataset serves as the foundation for our investigation on consumer reactions to postal marketing designed to promote financial goods. The data covers a diverse range of consumer characteristics, including demographic information, credit history, past campaign responses, and financial behavior measurements.

The main dataset comprises of records of individuals who were the focus of previous mail campaigns done by the fintech company. Every record includes multiple characteristics that depict the consumer's financial history and reaction to the campaign, such as whether they initiated a new credit card and their future credit behavior (delinquency status). The extensive dataset allows for a thorough examination of consumer behavior and reaction patterns, which is essential for optimizing future postal campaigns.

\subsection{Experimental Setup}

We used Python for data analysis and modeling because of its extensive ecosystem of modules and frameworks for data manipulation, statistical analysis, and machine learning. Pandas, Scikit-learn, and Matplotlib made Python perfect for complicated analytics and predictive models. The interactive development platform JupyterLab enabled exploratory data analysis and visualization in our research. JupyterLab's straightforward interface supports iterative Python code writing and execution, enabling rapid visualization and incremental construction of sophisticated data analysis algorithms. Coding, testing, and result interpretation are seamless in this environment because Python libraries are directly integrated.

\begin{table*}
\centering
\resizebox{\textwidth}{!}{
\begin{tabular}{|c|cccccccc|}
\hline
 &
  \multicolumn{8}{c|}{\textbf{Mail or Don't Mail Model Class}} \\ \cline{2-9} 
 &
  \multicolumn{4}{c|}{\textbf{Median  Imputation}} &
  \multicolumn{4}{c|}{\textbf{Custom bins Imputation}} \\ \cline{2-9} 
 &
  \multicolumn{2}{c|}{\textbf{quantile}} &
  \multicolumn{2}{c|}{\textbf{categorical}} &
  \multicolumn{2}{c|}{\textbf{quantile}} &
  \multicolumn{2}{c|}{\textbf{categorical}} \\ \cline{2-9} 
\multirow{-4}{*}{\textbf{Sampling}} &
  \multicolumn{1}{c|}{Logistic   Regression} &
  \multicolumn{1}{c|}{XgBoost} &
  \multicolumn{1}{c|}{Logistic Regression} &
  \multicolumn{1}{c|}{XgBoost} &
  \multicolumn{1}{c|}{Logistic Regression} &
  \multicolumn{1}{c|}{XgBoost} &
  \multicolumn{1}{c|}{Logistic Regression} &
  XgBoost \\ \hline
\textbf{Without   Oversampling} &
  \multicolumn{1}{c|}{\begin{tabular}[c]{@{}c@{}}Accuracy - 85.83\\      Precision - 86.91\\      Recall - 97.95\\      F-Score - 92.10\\      AUC - 81.24\end{tabular}} &
  \multicolumn{1}{c|}{\begin{tabular}[c]{@{}c@{}}Accuracy - 87.85\\      Precision - 89.13\\      Recall - 97.50\\      F-Score - 93.13\\      AUC - 83.17\end{tabular}} &
  \multicolumn{1}{c|}{\begin{tabular}[c]{@{}c@{}}Accuracy - 86.35\\      Precision - 87.49\\      Recall - 97.80\\      F-Score - 92.36\\      AUC - 82.44\end{tabular}} &
  \multicolumn{1}{c|}{\begin{tabular}[c]{@{}c@{}}Accuracy - 88.11\\      Precision - 89.26\\      Recall - 97.66\\      F-Score - 93.27\\      AUC - 83.75\end{tabular}} &
  \multicolumn{1}{c|}{\begin{tabular}[c]{@{}c@{}}Accuracy -   84.69\\      Precision - 84.99\\      Recall - 99.40\\      F-Score - 91.63\\      AUC - 72.91\end{tabular}} &
  \multicolumn{1}{c|}{\begin{tabular}[c]{@{}c@{}}Accuracy - 87.12\\      Precision - 87.84\\      Recall - 98.35\\      F-Score - 92.80\\      AUC - 76.36\end{tabular}} &
  \multicolumn{1}{c|}{\begin{tabular}[c]{@{}c@{}}Accuracy -   87.16\\      Precision - 88.43\\      Recall - 97.54\\      F-Score - 92.76\\      AUC - 85.23\end{tabular}} &
  {\color[HTML]{32CB00} \begin{tabular}[c]{@{}c@{}}Accuracy - 88.59\\      Precision - 89.82\\      Recall - 97.52\\      F-Score - 93.51\\      AUC - 85.90\end{tabular}} \\ \hline
\textbf{With   Random Oversampling} &
  \multicolumn{1}{c|}{\begin{tabular}[c]{@{}c@{}}Accuracy - 70.72\\      Precision - 94.03\\      Recall - 69.71\\      F-Score - 80.07\\      AUC - 81.14\end{tabular}} &
  \multicolumn{1}{c|}{\begin{tabular}[c]{@{}c@{}}Accuracy - 85.87\\      Precision - 90.61\\      Recall - 92.88\\      F-Score - 91.73\\      AUC - 82.57\end{tabular}} &
  \multicolumn{1}{c|}{\begin{tabular}[c]{@{}c@{}}Accuracy - 71.42\\      Precision - 94.23\\      Recall - 70.44\\      F-Score - 80.62\\      AUC - 82.43\end{tabular}} &
  \multicolumn{1}{c|}{{\color[HTML]{32CB00} \begin{tabular}[c]{@{}c@{}}Accuracy - 84.90\\      Precision - 91.12\\      Recall - 90.97\\      F-Score - 91.04\\      AUC - 83.04\end{tabular}}} &
  \multicolumn{1}{c|}{\begin{tabular}[c]{@{}c@{}}Accuracy -   65.34\\      Precision - 91.60\\      Recall - 64.86\\      F-Score - 75.94\\      AUC - 72.92\end{tabular}} &
  \multicolumn{1}{c|}{\begin{tabular}[c]{@{}c@{}}Accuracy - 83.90\\      Precision - 89.46\\      Recall - 91.72\\      F-Score - 90.58\\      AUC - 75.16\end{tabular}} &
  \multicolumn{1}{c|}{\begin{tabular}[c]{@{}c@{}}Accuracy -   73.53\\      Precision - 95.00\\      Recall - 72.43\\      F-Score - 82.11\\      AUC - 85.25\end{tabular}} &
  \begin{tabular}[c]{@{}c@{}}Accuracy - 85.00\\      Precision - 91.94\\      Recall - 90.12\\      F-Score - 91.02\\      AUC - 85.18\end{tabular} \\ \hline
\textbf{With   SMOTE Oversampling} &
  \multicolumn{1}{c|}{\begin{tabular}[c]{@{}c@{}}Accuracy - 84.99\\      Precision - 88.43\\      Recall - 94.57\\      F-Score - 91.40\\      AUC - 77.60\end{tabular}} &
  \multicolumn{1}{c|}{\begin{tabular}[c]{@{}c@{}}Accuracy - 87.61\\      Precision - 89.28\\      Recall - 96.96\\      F-Score - 92.96\\      AUC - 83.15\end{tabular}} &
  \multicolumn{1}{c|}{\begin{tabular}[c]{@{}c@{}}Accuracy - 85.82\\      Precision - 88.57\\      Recall - 95.52\\      F-Score - 91.91\\      AUC - 81.62\end{tabular}} &
  \multicolumn{1}{c|}{\begin{tabular}[c]{@{}c@{}}Accuracy - 86.54\\      Precision - 87.64\\      Recall - 97.84\\      F-Score - 92.46\\      AUC - 75.64\end{tabular}} &
  \multicolumn{1}{c|}{\begin{tabular}[c]{@{}c@{}}Accuracy - 83.65\\      Precision - 86.36\\      Recall - 95.73\\      F-Score - 90.80\\      AUC - 70.10\end{tabular}} &
  \multicolumn{1}{c|}{\begin{tabular}[c]{@{}c@{}}Accuracy - 86.54\\      Precision - 87.64\\      Recall - 97.84\\      F-Score - 92.46\\      AUC - 75.64\end{tabular}} &
  \multicolumn{1}{c|}{\begin{tabular}[c]{@{}c@{}}Accuracy - 86.57\\      Precision - 89.22\\      Recall - 95.63\\      F-Score - 92.31\\      AUC - 82.77\end{tabular}} &
  {\color[HTML]{32CB00} \begin{tabular}[c]{@{}c@{}}Accuracy - 88.38\\      Precision - 89.91\\      Recall - 97.13\\      F-Score - 93.38\\      AUC - 85.82\end{tabular}} \\ \hline
\end{tabular}
}
\caption{Experiment results of Mail or Don't Mail Model}
\label{tab:binary_class_model_results}
\end{table*}

\begin{table*}
\centering
\resizebox{\textwidth}{!}{
\begin{tabular}{|c|cccccccc|}
\hline
 &
  \multicolumn{8}{c|}{\textbf{Credit Model}} \\ \cline{2-9} 
 &
  \multicolumn{4}{c|}{\textbf{Median  Imputation}} &
  \multicolumn{4}{c|}{\textbf{Custom bins Imputation}} \\ \cline{2-9} 
 &
  \multicolumn{2}{c|}{\textbf{quantile}} &
  \multicolumn{2}{c|}{\textbf{categorical}} &
  \multicolumn{2}{c|}{\textbf{quantile}} &
  \multicolumn{2}{c|}{\textbf{categorical}} \\ \cline{2-9} 
\multirow{-4}{*}{\textbf{Sampling}} &
  \multicolumn{1}{c|}{\textbf{Logistic Regression}} &
  \multicolumn{1}{c|}{\textbf{XgBoost}} &
  \multicolumn{1}{c|}{\textbf{Logistic Regression}} &
  \multicolumn{1}{c|}{\textbf{XgBoost}} &
  \multicolumn{1}{c|}{\textbf{Logistic Regression}} &
  \multicolumn{1}{c|}{\textbf{XgBoost}} &
  \multicolumn{1}{c|}{\textbf{Logistic Regression}} &
  \textbf{XgBoost} \\ \hline
\textbf{Without   Oversampling} &
  \multicolumn{1}{c|}{\begin{tabular}[c]{@{}c@{}}Accuracy - 72.11\\      Precision - 64.76\\      Recall - 53.27\\      F-Score - 58.46\\      AUC - 77.22\end{tabular}} &
  \multicolumn{1}{c|}{\begin{tabular}[c]{@{}c@{}}Accuracy - 73.84\\      Precision - 66.71\\      Recall - 58.05\\      F-Score - 62.08\\      AUC - 79.18\end{tabular}} &
  \multicolumn{1}{c|}{\begin{tabular}[c]{@{}c@{}}Accuracy - 77.19\\      Precision - 74.54\\      Recall - 57.84\\      F-Score - 65.13\\      AUC - 85.74\end{tabular}} &
  \multicolumn{1}{c|}{{\color[HTML]{32CB00} \begin{tabular}[c]{@{}c@{}}Accuracy - 78.18\\      Precision - 73.00\\      Recall - 64.85\\      F-Score - 68.68\\      AUC - 86.59\end{tabular}}} &
  \multicolumn{1}{c|}{\begin{tabular}[c]{@{}c@{}}Accuracy - 71.38\\      Precision - 63.40\\      Recall - 52.75\\      F-Score - 57.59\\      AUC - 77.20\end{tabular}} &
  \multicolumn{1}{c|}{\begin{tabular}[c]{@{}c@{}}Accuracy - 73.94\\      Precision - 66.65\\      Recall - 58.74\\      F-Score - 62.45\\      AUC - 80.22\end{tabular}} &
  \multicolumn{1}{c|}{\begin{tabular}[c]{@{}c@{}}Accuracy - 77.12\\      Precision - 74.56\\      Recall - 57.49\\      F-Score - 64.92\\      AUC - 85.78\end{tabular}} &
  \begin{tabular}[c]{@{}c@{}}Accuracy - 78.10\\      Precision - 72.87\\      Recall - 64.72\\      F-Score - 68.55\\      AUC - 86.65\end{tabular} \\ \hline
\textbf{With   Random Oversampling} &
  \multicolumn{1}{c|}{\begin{tabular}[c]{@{}c@{}}Accuracy - 69.81\\      Precision - 57.13\\      Recall - 72.28\\      F-Score - 63.82\\      AUC - 77.22\end{tabular}} &
  \multicolumn{1}{c|}{\begin{tabular}[c]{@{}c@{}}Accuracy - 73.02\\      Precision - 63.47\\      Recall - 63.27\\      F-Score - 63.37\\      AUC - 78.93\end{tabular}} &
  \multicolumn{1}{c|}{\begin{tabular}[c]{@{}c@{}}Accuracy - 74.56\\      Precision - 61.93\\      Recall - 80.28\\      F-Score - 69.92\\      AUC - 85.74\end{tabular}} &
  \multicolumn{1}{c|}{\begin{tabular}[c]{@{}c@{}}Accuracy - 77.03\\      Precision - 67.70\\      Recall - 72.18\\      F-Score - 69.87\\      AUC - 86.43\end{tabular}} &
  \multicolumn{1}{c|}{\begin{tabular}[c]{@{}c@{}}Accuracy - 68.92\\      Precision - 55.93\\      Recall - 73.69\\      F-Score - 63.59\\      AUC - 77.19\end{tabular}} &
  \multicolumn{1}{c|}{\begin{tabular}[c]{@{}c@{}}Accuracy - 73.25\\      Precision - 63.74\\      Recall - 63.76\\      F-Score - 63.75\\      AUC - 80.04\end{tabular}} &
  \multicolumn{1}{c|}{\begin{tabular}[c]{@{}c@{}}Accuracy - 74.56\\      Precision - 61.79\\      Recall - 81.08\\      F-Score - 70.13\\      AUC - 85.79\end{tabular}} &
  {\color[HTML]{32CB00} \begin{tabular}[c]{@{}c@{}}Accuracy - 77.34\\      Precision - 68.13\\      Recall - 72.49\\      F-Score - 70.24\\      AUC - 86.54\end{tabular}} \\ \hline
\textbf{With   SMOTE Oversampling} &
  \multicolumn{1}{c|}{\begin{tabular}[c]{@{}c@{}}Accuracy - 69.41\\      Precision - 60.39\\      Recall - 49.29\\      F-Score - 54.28\\      AUC - 72.73\end{tabular}} &
  \multicolumn{1}{c|}{\begin{tabular}[c]{@{}c@{}}Accuracy - 73.63\\      Precision - 65.79\\      Recall - 59.40\\      F-Score - 62.43\\      AUC - 79.07\end{tabular}} &
  \multicolumn{1}{c|}{\begin{tabular}[c]{@{}c@{}}Accuracy - 76.15\\      Precision - 69.62\\      Recall - 62.52\\      F-Score - 65.88\\      AUC - 84.88\end{tabular}} &
  \multicolumn{1}{c|}{{\color[HTML]{32CB00} \begin{tabular}[c]{@{}c@{}}Accuracy - 77.84\\      Precision - 70.66\\      Recall - 68.29\\      F-Score - 69.46\\      AUC - 86.49\end{tabular}}} &
  \multicolumn{1}{c|}{\begin{tabular}[c]{@{}c@{}}Accuracy - 68.11\\      Precision - 58.29\\      Recall - 47.16\\      F-Score - 52.14\\      AUC - 71.91\end{tabular}} &
  \multicolumn{1}{c|}{\begin{tabular}[c]{@{}c@{}}Accuracy - 73.69\\      Precision - 65.78\\      Recall - 59.81\\      F-Score - 62.65\\      AUC - 80.13\end{tabular}} &
  \multicolumn{1}{c|}{\begin{tabular}[c]{@{}c@{}}Accuracy - 74.56\\      Precision - 61.79\\      Recall - 81.08\\      F-Score - 70.13\\      AUC - 85.79\end{tabular}} &
  \begin{tabular}[c]{@{}c@{}}Accuracy - 77.80\\      Precision - 70.52\\      Recall - 68.40\\      F-Score - 69.45\\      AUC - 86.60\end{tabular} \\ \hline
\end{tabular}
}
\caption{Experiment results of Credit Model}
\label{tab:credit_model_results}
\end{table*}

\begin{table*}
\centering
\resizebox{\textwidth}{!}{
\begin{tabular}{|c|cccccccc|}
\hline
 &
  \multicolumn{8}{c|}{\textbf{Responders Model}} \\ \cline{2-9} 
 &
  \multicolumn{4}{c|}{\textbf{Median  Imputation}} &
  \multicolumn{4}{c|}{\textbf{Custom bins Imputation}} \\ \cline{2-9} 
 &
  \multicolumn{2}{c|}{\textbf{quantile}} &
  \multicolumn{2}{c|}{\textbf{categorical}} &
  \multicolumn{2}{c|}{\textbf{quantile}} &
  \multicolumn{2}{c|}{\textbf{categorical}} \\ \cline{2-9} 
\multirow{-4}{*}{\textbf{Sampling}} &
  \multicolumn{1}{c|}{\textbf{Logistic Regression}} &
  \multicolumn{1}{c|}{\textbf{XgBoost}} &
  \multicolumn{1}{c|}{\textbf{Logistic Regression}} &
  \multicolumn{1}{c|}{\textbf{XgBoost}} &
  \multicolumn{1}{c|}{\textbf{Logistic Regression}} &
  \multicolumn{1}{c|}{\textbf{XgBoost}} &
  \multicolumn{1}{c|}{\textbf{Logistic Regression}} &
  \textbf{XgBoost} \\ \hline
\textbf{Without   Oversampling} &
  \multicolumn{1}{c|}{\begin{tabular}[c]{@{}c@{}}Accuracy - 75.59\\      Precision - 54.91\\      Recall - 10.18\\      F-Score - 17.17\\      AUC - 71.09\end{tabular}} &
  \multicolumn{1}{c|}{\begin{tabular}[c]{@{}c@{}}Accuracy - 76.24\\      Precision - 55.11\\      Recall - 24.17\\      F-Score - 33.61\\      AUC - 72.40\end{tabular}} &
  \multicolumn{1}{c|}{\begin{tabular}[c]{@{}c@{}}Accuracy - 76.04\\      Precision - 57.16\\      Recall - 14.58\\      F-Score - 23.23\\      AUC - 72.65\end{tabular}} &
  \multicolumn{1}{c|}{\begin{tabular}[c]{@{}c@{}}Accuracy - 81.31\\      Precision - 74.38\\      Recall - 37.94\\      F-Score - 50.25\\      AUC - 77.63\end{tabular}} &
  \multicolumn{1}{c|}{\begin{tabular}[c]{@{}c@{}}Accuracy - 80.31\\      Precision - 82.72\\      Recall - 93.28\\      F-Score - 87.68\\      AUC - 81.99\end{tabular}} &
  \multicolumn{1}{c|}{\begin{tabular}[c]{@{}c@{}}Accuracy - 81.31\\      Precision - 74.38\\      Recall - 37.94\\      F-Score - 50.25\\      AUC - 77.63\end{tabular}} &
  \multicolumn{1}{c|}{\begin{tabular}[c]{@{}c@{}}Accuracy - 80.29\\      Precision - 82.00\\      Recall - 94.53\\      F-Score - 87.82\\      AUC - 81.94\end{tabular}} &
  {\color[HTML]{32CB00} \begin{tabular}[c]{@{}c@{}}Accuracy - 83.60\\      Precision - 85.85\\      Recall - 93.60\\      F-Score - 89.56\\      AUC - 85.01\end{tabular}} \\ \hline
\textbf{With   Random Oversampling} &
  \multicolumn{1}{c|}{\begin{tabular}[c]{@{}c@{}}Accuracy - 63.05\\      Precision - 37.11\\      Recall - 70.02\\      F-Score - 48.51\\      AUC - 71.08\end{tabular}} &
  \multicolumn{1}{c|}{\begin{tabular}[c]{@{}c@{}}Accuracy - 68.11\\      Precision - 40.64\\      Recall - 61.29\\      F-Score - 48.88\\      AUC - 71.91\end{tabular}} &
  \multicolumn{1}{c|}{\begin{tabular}[c]{@{}c@{}}Accuracy - 64.60\\      Precision - 38.51\\      Recall - 70.99\\      F-Score - 49.93\\      AUC - 72.64\end{tabular}} &
  \multicolumn{1}{c|}{\begin{tabular}[c]{@{}c@{}}Accuracy - 78.01\\      Precision - 56.62\\      Recall - 49.60\\      F-Score - 52.88\\      AUC - 77.12\end{tabular}} &
  \multicolumn{1}{c|}{\begin{tabular}[c]{@{}c@{}}Accuracy - 72.44\\      Precision - 90.55\\      Recall - 70.70\\      F-Score - 79.40\\      AUC - 81.98\end{tabular}} &
  \multicolumn{1}{c|}{{\color[HTML]{32CB00} \begin{tabular}[c]{@{}c@{}}Accuracy - 82.16\\      Precision - 87.18\\      Recall - 89.40\\      F-Score - 88.28\\      AUC - 84.91\end{tabular}}} &
  \multicolumn{1}{c|}{\begin{tabular}[c]{@{}c@{}}Accuracy - 71.38\\      Precision - 90.09\\      Recall - 69.55\\      F-Score - 78.50\\      AUC - 81.93\end{tabular}} &
  \begin{tabular}[c]{@{}c@{}}Accuracy - 82.04\\      Precision - 86.79\\      Recall - 89.74\\      F-Score - 88.24\\      AUC - 84.42\end{tabular} \\ \hline
\textbf{With   SMOTE Oversampling} &
  \multicolumn{1}{c|}{\begin{tabular}[c]{@{}c@{}}Accuracy - 66.97\\      Precision - 38.36\\      Recall - 54.12\\      F-Score - 44.90\\      AUC - 68.52\end{tabular}} &
  \multicolumn{1}{c|}{\begin{tabular}[c]{@{}c@{}}Accuracy - 71.91\\      Precision - 44.11\\      Recall - 48.36\\      F-Score - 46.41\\      AUC - 71.28\end{tabular}} &
  \multicolumn{1}{c|}{\begin{tabular}[c]{@{}c@{}}Accuracy - 75.01\\      Precision - 49.49\\      Recall - 23.44\\      F-Score - 31.81\\      AUC - 70.06\end{tabular}} &
  \multicolumn{1}{c|}{\begin{tabular}[c]{@{}c@{}}Accuracy - 83.60\\      Precision - 85.85\\      Recall - 93.60\\      F-Score - 89.56\\      AUC - 85.01\end{tabular}} &
  \multicolumn{1}{c|}{\begin{tabular}[c]{@{}c@{}}Accuracy - 79.38\\      Precision - 83.94\\      Recall - 89.72\\      F-Score - 86.73\\      AUC - 80.97\end{tabular}} &
  \multicolumn{1}{c|}{\begin{tabular}[c]{@{}c@{}}Accuracy - 83.60\\      Precision - 85.85\\      Recall - 93.60\\      F-Score - 89.56\\      AUC - 85.01\end{tabular}} &
  \multicolumn{1}{c|}{\begin{tabular}[c]{@{}c@{}}Accuracy - 79.74\\      Precision - 82.83\\      Recall - 92.14\\      F-Score - 87.22\\      AUC - 80.94\end{tabular}} &
  {\color[HTML]{32CB00} \begin{tabular}[c]{@{}c@{}}Accuracy - 83.64\\      Precision - 85.30\\      Recall - 94.52\\      F-Score - 89.67\\      AUC - 84.63\end{tabular}} \\ \hline
\end{tabular}
}
\caption{Experiment results of Responders Model}
\label{tab:responders_model_results}
\end{table*}

\subsection{Results}

The study we conducted involved three separate prediction models: Mail or Don't Mail, Credit, and Responders. To improve the performance of these models, we utilized several data preparation approaches. We conducted experiments using two imputation methods: custom bin imputation, which preserves the original data distribution by binning and assigning mean values based on similar excellent rates, and mean imputation, which replaces missing data with the global mean but may be influenced by outliers.

We utilized two binning strategies, namely quantile binning and categorical binning, for each imputation method. Quantile binning divides the data into bins of equal size, making it suitable for skewed distributions. On the other hand, categorical binning classifies the data based on intrinsic category similarities. In order to tackle the issue of class imbalance, we used random oversampling and SMOTE (Synthetic Minority Over-sampling Technique) into our research. Subsequently, logistic regression and XGBoost methods were employed to model each variety of the dataset. These trials yielded valuable information about how various data preprocessing and sampling procedures affect the predicted accuracy and resilience of each model. They demonstrated the subtle effects of each technique on our ability to correctly forecast customer behavior and creditworthiness.

\begin{figure*}
   \begin{minipage}{0.45\textwidth}
     \centering
     \includegraphics[width=0.8\linewidth,height=1.4in]{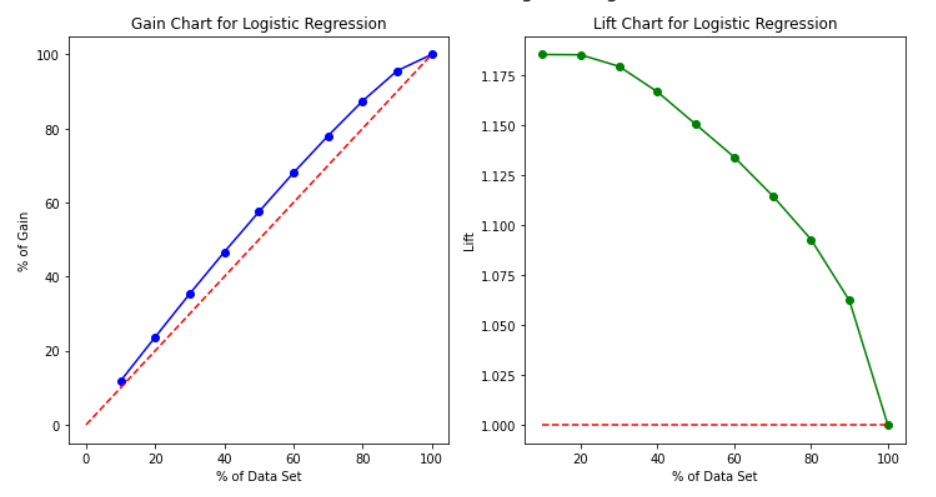}
    \caption{Gain/Lift Chart of Best Performing Logistic Regression Model}
    \label{fig:logistic_gain_lift}
   \end{minipage}\hfill
   \begin{minipage}{0.4\textwidth}
     \centering
    \includegraphics[width=0.8\linewidth,height=1.4in]{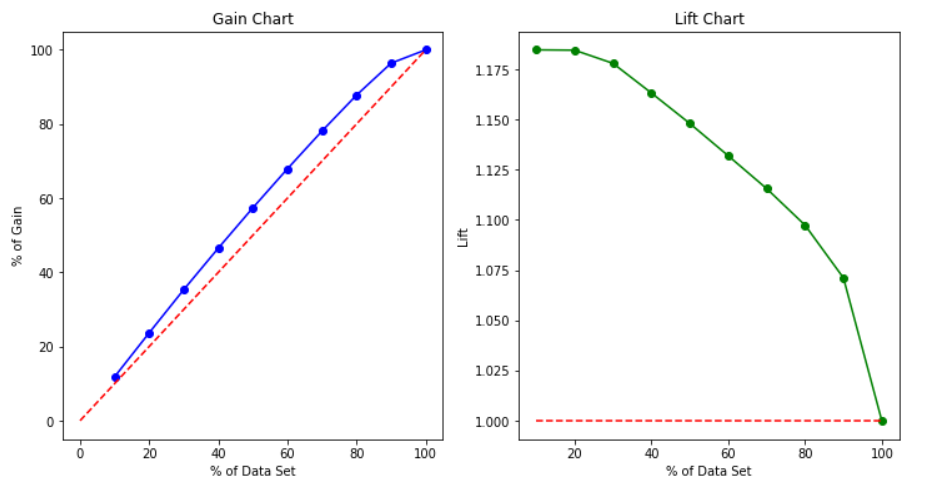}
    \caption{Gain/Lift Chart of Best Performing XgBoost Model}
    \label{fig:xgboost_gain_lift}
   \end{minipage}
\vspace{-0.1in}
\end{figure*}

The experimental results summarized in Table \ref{tab:binary_class_model_results} provide a comprehensive evaluation of the Mail or Don't Mail model under different data preprocessing conditions. Custom bin imputation, particularly when combined with categorical binning, demonstrated superior performance across key metrics like accuracy (88.59\%) and AUC(85.90\%), especially with the XGBoost algorithm. This suggests that aligning imputation closely with the data's distribution notably enhances model performance. Furthermore, the use of SMOTE oversampling was effective in improving recall rates, indicating its utility in mitigating class imbalance by synthesizing new examples. Overall, XGBoost outperformed logistic regression, showcasing its strength in handling complex, non-linear relationships, which is crucial for optimizing marketing strategies in a highly competitive fintech environment.

Similarly, Table \ref{tab:credit_model_results} displays the results of all the experiments conducted on the Credit model. The best-performing model for each combination of imputation and binning technique across different sampling methods is highlighted in green. In the Credit model, the XGBoost algorithm paired with median imputation and categorical binning emerged as the standout, achieving an accuracy of 78.18 and an AUC of 86.59. This combination's superior performance underscores its efficacy in accurately predicting outcomes within the credit assessment context.

Finally, Table \ref{tab:responders_model_results} presents the results from all experiments conducted on the Responders model. The best-performing model for each combination of imputation and binning technique across different sampling methods, is highlighted in green. Within the Responders model, the XGBoost model with custom imputation and categorical binning delivered the most impressive results, achieving an accuracy of 83.60 and an AUC of 85.01. This demonstrates its effectiveness in accurately identifying likely responders, optimizing targeted strategies for mail campaigns.

Figure \ref{fig:logistic_gain_lift} presents the gain/lift chart of the best-performing logistic regression model, illustrating its effectiveness in distinguishing between classes within the dataset. Similarly, Figure \ref{fig:xgboost_gain_lift} showcases the gain/lift chart for the best-performed XGBoost model, highlighting its superior predictive capabilities and how effectively it leverages the underlying patterns in the data to optimize outcomes. These visual representations provide a clear comparative insight into the performance enhancements achieved through advanced modeling techniques.

\section{Conclusion} \label{sec: Conclusion}

During the course of our experimentation, the XGBoost model often achieved better results than logistic regression, especially in situations that involved categorical binning and bespoke imputation. The consistent occurrence of this pattern was observed in many datasets and preprocessing procedures, which emphasized the exceptional capability of XGBoost in managing intricate and non-linear interactions within the data. The resilience and adaptability of XGBoost in these contexts highlight its usefulness, making it a particularly beneficial tool for predictive modeling in financial technology applications. The correct prediction of customer behavior and creditworthiness greatly improves the strategic implementation of marketing efforts, highlighting the potential of modern machine learning approaches to inform corporate decisions and promote operational efficiencies.

\bibliographystyle{IEEEtran}
\bibliography{refes}
\end{document}